\documentclass[10pt, a4paper]{article}
\usepackage[final]{lrec2026} 

\usepackage{multirow}
\usepackage{amssymb}
\usepackage{booktabs} 
\usepackage{fontspec}
\usepackage{polyglossia}
\setmainlanguage{english}
\setotherlanguage{arabic}
\newfontfamily\arabicfont[Script=Arabic,Scale=1.1]{Amiri-Regular.ttf}

\title{StanceNakba Shared Task: Actor and Topic-Aware Stance Detection in Public Discourse}

\name{Kholoud K. Aldous$^1$, Md Rafiul Biswas$^2$, Mabrouka Bessghaier$^1$,  \\[2pt]
\fontsize{12}{11}\selectfont \textbf{Shimaa Ibrahim$^1$, Kais Attia}, \textbf{Wajdi Zaghouani$^1$}}
\address{$^1$Northwestern University in Qatar, $^2$Hamad Bin Khalifa University\\
        mbiswas@hbku.edu.qa,\\
         \{kholoud.aldous, mabrouka.bessghaier, shimaa.ibrahim, wajdi.zaghouani\}@northwestern.edu\\
         Kais.attia.w@gmail.com}

\abstract{
We present StanceNakba 2026, a shared task on stance detection in polarized social media discourse related to the Palestinian-Israeli conflict, organized as part of Nakba-NLP 2026 at LREC-COLING 2026. The task introduces two subtasks: Subtask~A (Actor-Level Stance Detection), which classifies English social media posts as Pro-Palestine, Pro-Israel, or Neutral; and Subtask~B (Cross-Topic Stance Detection), which identifies Favor, Against, or Neither stances in Arabic posts toward two conflict-related topics, normalization with Israel and refugee presence in Jordan. The task is grounded in an annotated dataset of 2,606 social media posts. A total of 7 teams participated in Subtask~A and 6 teams in Subtask~B. Participating systems primarily fine-tuned Arabic and multilingual transformer-based models, including MARBERT, AraBERT, and DeBERTa-v3 variants, with several teams employing cross-validation, ensemble methods, and topic-conditioned architectures. The best-performing systems achieved a Macro F1 of 0.9620 on Subtask~A and 0.8724 on Subtask~B, demonstrating that transformer-based approaches are highly effective for conflict-domain stance detection while highlighting persistent challenges in cross-topic generalization and neutral class prediction.
\\ \newline  }

\begin{document}

\maketitleabstract

\section{Introduction}
Social media platforms now serve as key spaces for political discussion, especially when it comes to deeply polarized issues like the Palestinian–Israeli conflict. The volume, diversity, and emotional intensity of online discussions on this topic present significant challenges for natural language processing (NLP) systems that aim to understand public opinion and political orientation. Stance detection is the task of automatically identifying whether an author is in favor of, against, or neutral toward a given target~\cite{mohammad2016semeval}. It offers a principled framework for analyzing such discourse. However, existing resources and shared tasks have largely focused on English and on politically neutral topics, leaving conflict-related, multilingual, and polarized discourse substantially underexplored.

To address this gap, we present the \textbf{StanceNakba 2026 Shared Task}, organized as part of Nakba-NLP 2026: The 2nd International Workshop on Nakba Narratives as Language Resources~\cite{JHE26}, co-located with LREC-COLING 2026. StanceNakba 2026 introduces two subtasks to stance detection that distinguish between actor-level political alignments and cross-topic stance patterns in social media posts related to the Palestinian-Israeli conflict and associated regional issues. The task is built on an annotated dataset of 2,606 social media posts in English and Arabic.

The shared task contains two subtasks. \textbf{Subtask A} (Actor-Level Stance Detection) is an English-language task that requires systems to identify whether the author of a social media post expresses a \textit{Pro-Palestine}, \textit{Pro-Israel}, or \textit{Neutral} orientation in their general position toward the Palestinian-Israeli conflict. \textbf{Subtask B} (Cross-Topic Stance Detection) is an Arabic-language task that requires systems to detect \textit{Favor}, \textit{Against}, or \textit{Neither} stances toward two specific conflict-related topics: normalization with Israel and refugee presence in Jordan. These subtasks enable the investigation of fundamental questions at the intersection of NLP and political discourse analysis: How do general political alignments relate to positions on specific policy issues? Can models learn generalizable stance representations that transfer across different topics within the same conflict domain?

The task attracted broad participation from the research community. A total of 7 teams submitted systems for Subtask~A and 6 teams for Subtask~B, with two teams participating in both. Participating systems primarily relied on fine-tuned Arabic and multilingual transformer-based models, including MARBERT~\cite{abdul2021arbert}, AraBERT~\cite{antoun2020arabert}, and DeBERTa-v3~\cite{he2021debertav3} variants, with the best system achieving a Macro F1 of 0.9620 on Subtask~A and 0.8724 on Subtask~B. Other techniques such as cross-validation, ensemble methods, Natural Language Inference (NLI) reformulation, and topic-conditioned architectures were explored, demonstrating the richness of approaches that this task inspired.

The remainder of this paper is organized as follows. Section~\ref{sec:related} reviews related work on stance detection and Arabic NLP. Section~\ref{sec:results-a} presents the results and system overview for Subtask~A, followed by Section~\ref{sec:results-b} for Subtask~B. Section~\ref{sec:conclusion} concludes with a discussion of findings and directions for future work. Section~\ref{sec:limitations} discusses the limitations of the current task and dataset. Finally, Section~\ref{sec:ethical} provides details about the ethical considerations and dataset availability.



\section{Related Work}
\label{sec:related}
Stance detection is the task of automatically determining whether the author of a text is in favor of, against, or neutral toward a given target~\cite{mohammad2016semeval}. 
SemEval-2016 Task~6~\cite{mohammad2016semeval} marked a turning point by introducing a widely used benchmark for stance detection in English tweets, covering targets such as political figures, social movements, and ideological positions.
Since then, stance detection has been studied across a broad range of social and political domains, with surveys by \citet{aldayel2021stance} documenting the rapid growth of the field and the shift from traditional feature-based methods toward transformer-based architectures.
Transformer-based models have become the dominant paradigm in stance detection. Pre-trained language models such as BERT~\cite{devlin2019bert} have been extensively fine-tuned for stance classification, consistently outperforming earlier approaches that relied on lexical features, n-grams, and handcrafted representations. For the Arabic language, dedicated pre-trained models have proven particularly important due to Arabic's morphological richness, dialectal diversity, and diglossia. AraBERT~\cite{antoun2020arabert} established strong baselines for Modern Standard Arabic NLP tasks, while ARBERT and MARBERT~\cite{abdul2021arbert} extended coverage to dialectal and social media Arabic by pre-training on large-scale Twitter corpora. 
Arabic stance detection systems now widely rely on these models as their core architecture, as seen in the approaches used by participants in the StanceNakba~2026 shared task.

Arabic stance detection has received growing attention in recent years, though resources remain limited compared to English. The Mawqif dataset~\cite{alturayeif2022mawqif} introduced the first Arabic target-specific stance corpus, comprising over 4,000 tweets annotated for stance, sentiment, and sarcasm across multiple controversial topics. Complementing this, \citet{article1} introduced a cross-domain, multi-dialectal Arabic stance corpus covering four Arab regions and multiple dialect groups, with over 4,500 annotated sentences balanced across MSA and regional dialects; their AraBERT-based system achieved strong performance and notably outperformed Mawqif-trained models on the neutral stance class, highlighting the importance of class balance in Arabic stance resources. Building on these resources, StanceEval~2024~\cite{alturayeif2024stanceeval} organized the first shared task dedicated to Arabic stance detection, hosted at ArabicNLP~2024. Participating systems mainly fine-tuned Arabic BERT variants, with ensemble methods and multi-task learning emerging as effective strategies for handling class imbalance and dialectal variation. These efforts demonstrated both the feasibility of Arabic stance detection at scale and the challenges that remain, particularly for underrepresented and neutral stance classes.

The StanceNakba~2026 shared task builds upon these foundations while introducing unique challenges tied to the highly polarized nature of discourse surrounding the Palestinian-Israeli conflict. Prior NLP work on conflict-related discourse has examined stance and opinion on related topics, including studies of Twitter polarization around the Israel-Palestine conflict~\cite{Imtiaz2022TakingSP} and cross-conflict stance correlations~\cite{tao2024eyes}. StanceNakba~2026 is organized as part of the Nakba-NLP~2026 workshop, the second edition of an initiative dedicated to applying NLP tools to the documentation and understanding of Nakba narratives. 
The shared task distinguishes itself from prior Arabic stance work through its dual-framework design: Subtask~A addresses actor-level political alignment in English, classifying authors as Pro-Palestine, Pro-Israel, or Neutral, while Subtask~B targets cross-topic Arabic stance detection toward specific conflict-related issues, namely normalization with Israel and refugee presence in Jordan. 

Recent work has increasingly emphasized the importance of modeling stance and narrative framing in politically sensitive and conflict-driven contexts. In the Arabic NLP domain, several efforts have focused on capturing polarization, ideology, and media narratives through dedicated datasets and shared tasks. For instance, the FIGNEWS shared task \cite{zaghouani2024fignews} introduced a benchmark for analyzing news media narratives, highlighting the role of framing and bias in shaping public discourse.

Complementary work has explored conflict-related and politically charged discourse on social media platforms. \cite{shestakov2024sheikhjarrah} presented a dataset analyzing the digital framing of the Sheikh Jarrah evictions, providing insights into how conflict narratives are constructed and propagated online. Similarly, \cite{alheraki2025hijab} examined polarization and misogynistic discourse in Arabic Twitter conversations, further demonstrating the challenges of modeling stance in emotionally charged environments.

Beyond stance-specific datasets, related efforts in hate speech and propaganda detection have contributed to understanding polarized language and ideological positioning. Resources such as multi-label hate speech corpora \cite{zaghouani2024hate} and shared tasks on propaganda and subjectivity detection \cite{hasanain2024checkthat} provide complementary perspectives on how stance, bias, and persuasion interact in online discourse.

These works collectively underscore the need for domain-specific, conflict-aware resources and evaluation frameworks. The StanceNakba shared task builds on this line of research by introducing actor-level and cross-topic stance detection in the context of the Palestinian-Israeli conflict, extending prior work toward more fine-grained and context-sensitive modeling of political stance.



\section{Subtask A: Actor-Level Stance Detection}
\label{sec:results-a}
Subtask~A frames conflict stance detection at the \textit{actor level}: given a social media post authored by a single user, the model must infer the author's overarching political orientation toward the Palestinian-Israeli conflict. Rather than analyzing individual argumentative moves or claim-specific positions, the task requires models to aggregate signals across the full text and assign a single author-level stance label. This formulation is motivated by the observation that political orientation in polarized discourse is often expressed indirectly through framing choices (which actors are foregrounded, which actions are lexicalized as aggression versus defense, which populations are centered) rather than through explicit opinion markers alone.


\subsection{Dataset}
The dataset consists of 1,401 English-language posts collected from X (formerly Twitter), all related to the Palestinian-Israeli conflict. Posts were retrieved via the Meltwater media intelligence platform using targeted keyword queries: \textit{``I stand with Palestine''} or \textit{``I stand with Gaza''} for the Pro-Palestine class, \textit{``I stand with Israel''} for the Pro-Israel class, and \textit{``Israel AND Palestine OR Gaza''} for the Neutral class. Labels were assigned based on the query that retrieved each post, yielding a balanced dataset of 467 samples per class (33.3\% each).

To ensure data quality, a multi-step preprocessing pipeline was applied. First, stance-signaling seed keywords (e.g., ``I stand with Palestine'') were stripped to prevent label leakage, along with Twitter mentions, URLs, hashtags, emojis, and special characters, retaining only alphanumeric text and basic punctuation. Posts were then filtered for meaningful content, requiring a minimum of 30 characters and at least five dictionary words of three or more letters, thereby excluding near-empty or uninformative entries. Finally, case-insensitive exact deduplication was performed by comparing lowercase-normalized versions of the cleaned texts, retaining only the first occurrence of each unique post.

\textbf{Data splits:} The dataset is partitioned into training (980 samples, 70\%), development (210 samples, 15\%), and test (211 samples, 15\%) splits, each preserving the balanced class distribution. Table~\ref{tab:subtaskA-data} summarizes the dataset statistics.

\begin{table}[t]
\centering
\small
\caption{Subtask A dataset statistics.}
\label{tab:subtaskA-data}
\begin{tabular}{lcccc}
\toprule
\textbf{Label} & \textbf{Train} & \textbf{Dev} & \textbf{Test} & \textbf{Total} \\
\midrule
Pro-Palestine & 327 & 70 & 70 & 467 \\
Pro-Israel    & 327 & 70 & 70 & 467 \\
Neutral       & 326 & 70 & 71 & 467 \\
\midrule
\textbf{Total} & \textbf{980} & \textbf{210} & \textbf{211} & \textbf{1,401} \\
\bottomrule
\end{tabular}
\end{table}

\textbf{Dataset Examples: }The following examples illustrate the range of expression captured by each label:

\begin{itemize}
    \item \textbf{Pro-Palestine:} \textit{``The systematic displacement of Palestinian families from their ancestral homes represents a clear violation of international law and the right of return.''}

    \item \textbf{Pro-Israel:} \textit{``Israel's defensive measures are necessary responses to existential threats, ensuring the safety of its citizens against terrorism.''}

    \item \textbf{Neutral:} \textit{``The conflict involves competing territorial claims, with both populations having deep historical connections to the region.''}
\end{itemize}



    
    

\subsection{Setup and Evaluation}
\label{sec:taskSetup}
The task was organized into two phases:
\begin{itemize}
    \item \textbf{Development phase}: we released the training and development subsets, and participants submitted runs on the development set through a competition on CodaBench.\footnote{\url{https://www.codabench.org/}}
    \item \textbf{Test phase}: we released the official test subset, and participants were given a few days to submit their final predictions through the same CodaBench competition. Only the latest submission from each team was considered official and was used for the final team ranking.
\end{itemize}

\noindent\textbf{Measures:} Subtask~A is framed as a three-class classification problem over English social media posts, where participants are asked to build a single unified model that maps an input post to one of three actor-level stance labels: \textsc{Pro-Palestine}, \textsc{Pro-Israel}, or \textsc{Neutral}. We measure the performance of the participating systems using \textbf{Macro F1}, which computes the unweighted mean of per-class F1 scores, weighting each class equally and thus penalizing poor performance on any individual label, including the minority-leaning \textsc{Neutral} class. Systems are evaluated on a held-out test set, and participants were not permitted to use the test set for model selection or hyperparameter tuning. In addition to Macro F1, Accuracy, Precision, and Recall are also reported on the leaderboard for reference.



\subsection{Results and Overview of the Systems}
A total of 12 teams submitted runs during the evaluation phase of Subtask~A (Actor-Level Stance Detection), of which 7 teams submitted system descriptions and 6 teams submitted system papers. In Table~\ref{tab:subtaskA-systems}, we provide an overview of the participating systems for which a description was submitted. In Table~\ref{tab:subtaskA-results}, we report the official results for those teams, ranked by Macro F1.


As shown in Table~\ref{tab:subtaskA-systems}, fine-tuning pre-trained transformer-based models is the dominant approach among participating teams, with BERT-based architectures being the most common backbone. Several teams additionally employed cross-validation, ensemble methods, and hyperparameter optimization.

\begin{table*}[t]
\centering
\small
\caption{Subtask A: Overview of the participating systems. DL: Deep Learning, ML: Classic Machine Learning.}
\label{tab:subtaskA-systems}
\resizebox{\textwidth}{!}{%
\begin{tabular}{lcccccccccccc}
\toprule
\multirow{2}{*}{\textbf{Team}} &
\multicolumn{6}{c}{\textbf{Transformer}} &
\multicolumn{1}{c}{\textbf{DL}} &
\multicolumn{5}{c}{\textbf{Misc.}} \\
\cmidrule(lr){2-7}\cmidrule(lr){8-8}\cmidrule(lr){9-13}
 & \rotatebox{90}{BERT-base} & \rotatebox{90}{XLM-RoBERTa} & \rotatebox{90}{ARBERT/MARBERT} & \rotatebox{90}{DeBERTa-v3} & \rotatebox{90}{mDeBERTa-v3} & \rotatebox{90}{KE-MLM BERT} & \rotatebox{90}{CNN/MoE} & \rotatebox{90}{Data Augmentation} & \rotatebox{90}{Preprocessing} & \rotatebox{90}{Hyperparameter Tuning} & \rotatebox{90}{Cross-validation} & \rotatebox{90}{Ensemble Methods} \\
\midrule
Shroukgbr~\cite{shroukgbr_2026}      & \checkmark & \checkmark & \checkmark &            &            &            &            & \checkmark & \checkmark & \checkmark & \checkmark & \checkmark \\
Yafa~\cite{yafa_2026}      &            &            & \checkmark &            &            &            &            &            &            & \checkmark &            &            \\
KUET~\cite{teamkuet_2026}     & \checkmark &            &            &            &            &            & \checkmark &            &            & \checkmark & \checkmark & \checkmark \\
KvochurHegel~\cite{kvochurhegel_2026}   &            &            &            & \checkmark &            &            &            &            &            & \checkmark &            &            \\
Viva\_Palestine~\cite{vivapalestine_2026}& \checkmark &            &            &            &            &            &            &            &            & \checkmark &            &            \\
HeatWave       &            &            &            & \checkmark &            & \checkmark &            &            &            & \checkmark & \checkmark & \checkmark \\
The Blackwell Collective~\cite{blackwell_2026} &  &            &            &            & \checkmark &            &            &            &            & \checkmark &            &            \\
\bottomrule
\end{tabular}%
}
\end{table*}

\begin{table}[t]
\centering
\small
\caption{Official results for Subtask A, ranked by Macro F1 score. Leaderboard ranks are shown alongside overall submission ranks (in parentheses). Of the 12 total test submissions, only 7 teams provided a model description and are included here.}
\label{tab:subtaskA-results}
\resizebox{\linewidth}{!}{%
\begin{tabular}{lcc}
\toprule
\textbf{Team} & \textbf{Rank} & \textbf{Macro F1} \\
\midrule
Shroukgbr                & 1 (\#1) & 0.9620 \\
Yafa                & 2 (\#2) & 0.9525 \\
KUET               & 3 (\#3) & 0.9426 \\
KvochurHegel              & 4 (\#4) & 0.9384 \\
Viva\_Palestine          & 5 (\#6) & 0.9190 \\
HeatWave                 & 6 (\#7) & 0.8804 \\
The Blackwell Collective & 7 (\#11)& 0.7451 \\
\bottomrule
\end{tabular}
}
\end{table}

Team \textbf{Shroukgbr}~\cite{shroukgbr_2026} achieved the best performance with a Macro F1 of 0.9620, ranking first on the leaderboard. They fine-tuned a single BERT-based transformer model initialized from a publicly available pretrained checkpoint. Preprocessing steps included removing URLs and user mentions, normalizing repeated characters, and truncating inputs to 256 tokens. To address class imbalance, they applied weighted cross-entropy loss. The final configuration used stratified 5-fold cross-validation, an AdamW optimizer with a learning rate of 2e-5, batch size of 16, and 4 training epochs with early stopping.

Team \textbf{Yafa}~\cite{yafa_2026} ranked second with a Macro F1 of 0.9525. They fine-tuned MARBERT and ARBERT using a cross-lingual approach.  A customized cross-entropy loss with label smoothing and increased attention dropout (0.2) were applied to mitigate overconfidence and overfitting. Preprocessing included lowercase normalization, URL removal, user mention normalization, character repetition removal, whitespace normalization, hashtag normalization, and connected word splitting. The best results were obtained with a learning rate of $2\times10^{-5}$, weight decay of 0.1, batch size of 4, maximum sequence length of 128, and 8 epochs with early stopping (patience = 2) and an AdamW optimizer.

Team \textbf{KUET}~\cite{teamkuet_2026} ranked third with a Macro F1 of 0.9426. They fine-tuned a BERT-based Mixture-of-Experts (MoE) stance classification model built on top of \texttt{bert-base-uncased}. The architecture incorporated cue-word masking and contrast-marker amplification within specialized expert modules, along with CNN-based feature extraction with kernel sizes $\{2, 3, 4, 5\}$. They employed 10-fold stratified cross-validation with a weighted ensemble at inference time, where each fold model's contribution was proportional to its validation Macro F1. Label smoothing ($\epsilon=0.25$) was applied to reduce overconfidence.

Team \textbf{KvochurHegel}~\cite{kvochurhegel_2026} ranked fourth with a Macro F1 of 0.9384. They formulated stance detection as a Natural Language Inference (NLI) task, initializing a Cross-Encoder from the \texttt{DeBERTa-v3-base-mnli-fever-anli} checkpoint. Each input text was evaluated against three class-specific engineered hypotheses designed to encode ideological markers. Label smoothing (factor = 0.2) and R-Drop consistency regularization ($\alpha = 4.0$) were applied to mitigate label noise. The model used AdamW with a learning rate of 2e-5, batch size of 16, and early stopping with patience of 2 epochs.

Team \textbf{Viva\_Palestine}~\cite{vivapalestine_2026} ranked fifth with a Macro F1 of 0.9190. They pretrainined \texttt{bert-base-uncased} and fine-tuned it by merging training and validation data for the test phase, training for 20 epochs with early stopping (stopping at epoch 9), a maximum token length of 256, and a learning rate of $10^{-5}$.

Team \textbf{HeatWave}\footnote{This team didn't submit their system paper} ranked sixth with a Macro F1 of 0.8804. They experimented with multiple transformer-based models and large language models (LLMs), including \texttt{microsoft/deberta-v3-large}, \texttt{DeBERTa-v3-large-mnli-fever-anli-ling-wanli}, and the election-domain stance model \texttt{bert-election2020-twitter-stance-biden-KE-MLM}, the latter achieving the best results upon fine-tuning. They also explored zero-shot prompting with Qwen2.5 and DeepSeek-R1 models, though these yielded lower F1 scores (0.2--0.6). N-fold training with soft ensemble inference was used for the final submission.

Team \textbf{The Blackwell Collective}~\cite{blackwell_2026} ranked seventh with a Macro F1 of 0.7451. They fine-tuned \texttt{microsoft/mdeberta-v3-base} using a Statement Tuning paradigm: inputs were reformatted as cloze templates and the \texttt{[MASK]} token's hidden state was contrasted against three learnable class prototypes via a softmax head. Three auxiliary modules were incorporated: Prototype Contrastive Learning (PCL) with $K=3$ global prototypes at temperature $\tau=0.1$, Topic-Conditioned Layer Normalization (T-CLN) conditioning feature distributions on topic identity, and R-Drop consistency regularization via symmetric KL divergence. Training followed a two-stage schedule: backbone frozen for 2 epochs, then full fine-tuning for 8 epochs, using AdamW with a learning rate of $10^{-5}$ for the backbone and $10^{-4}$ for the heads.





    
    

\section{Subtask B: Cross-Topic Stance Detection}
\label{sec:results-b}  
Subtask B focuses on cross-topic stance detection, where systems are required to determine the stance expressed in a text toward a given target topic. In the context of the StanceNakba Shared Task, this subtask centers on politically sensitive discourse related to the Palestinian cause and regional political dynamics.

\subsection{Dataset}

We used a subset of the MARASTA dataset \cite{article1} for this subtask. MARASTA is a cross-domain, multidialectal Arabic stance corpus designed to support stance detection across multiple dialectal regions of the Arab world. The dataset covers four major dialectal regions, namely Maghreb, Egypt, the Levant, and the Gulf, representing the primary dialect groups in Arabic.

\paragraph{Dataset Overview}

The full MARASTA corpus contains more than 4,500 sentences annotated with stance toward eight controversial topics distributed across the four Arab regions. Each sentence is labeled according to its stance toward a given topic using three categories: pro (favor), against, or neutral. The dataset was constructed with careful balancing strategies. Each topic contains approximately 500-700 sentences, with a relatively even distribution across the three stance classes. In addition, the corpus maintains a balance between MSA and dialectal Arabic, with roughly half of the sentences written in MSA and the remaining half in the dialect associated with the topic’s region.

\paragraph{Data Collection}

The data were collected from online platforms such as Twitter and YouTube, where discussions on controversial socio-political topics are common. The collected content could be either current or historically relevant within the past 15 years. For each region (Maghreb, Egypt, the Levant, and the Gulf), the two most highly discussed topics were selected. Candidate sentences were retrieved using seed keywords associated with each topic. To do so, the authors used Python scripts leveraging the platforms’ APIs to retrieve tweets and YouTube comments containing the selected keywords. The collected sentences were then manually filtered to ensure they were written in Arabic (MSA or dialect), grammatically correct, relevant to the topic, and expressed a stance (pro, against) or a neutral position, allowing the dataset to capture naturally occurring stance expressions across different dialects and communication styles.

\paragraph{Data Annotation} The annotation process followed a multi-stage quality control procedure. Each sentence was independently annotated by two annotators to determine its stance with respect to the associated topic. In cases of disagreement, the instance was reviewed by a third annotator who resolved the conflict and determined the final label. The resulting annotations demonstrate substantial inter-annotator agreement, with an overall Cohen’s Kappa score of 0.84 and regional agreement scores ranging from 0.80 to 0.89, indicating high annotation reliability.\\

\paragraph{Topic Selection for the StanceNakba Shared Task} 

Since the StanceNakba Shared Task focuses on political discourse and narratives related to the Palestinian cause and regional political dynamics, we selected the two topics from the MARASTA corpus that are most relevant to this theme: ``Normalization with Israel'' and ``Refugee/Immigrant Presence in Jordan''. The dialectal content associated with these topics reflects two distinct Arabic regional varieties: the ``Normalization with Israel'' topic contains Gulf Arabic and MSA, while the ``Refugee/Immigrant Presence in Jordan'' topic contains Levantine Arabic (Jordanian/Palestinian) and MSA. These topics capture key socio-political debates in the Arab world that are closely connected to discussions surrounding Palestine, regional diplomacy, and displacement in the Middle East. The main characteristics of the subset used in this shared task are summarized below:

\begin{itemize}
    \item \textbf{Source:} Arabic social media posts (X) discussing Palestinian-Israeli conflict-related issues.
    \item \textbf{Total Size:} 1,205 annotated samples across two topics
    \item \textbf{Topics Covered:}
\end{itemize}

\textbf{Topic 1:\textarabic{ التطبيع مع إسرائيل} (Normalization with Israel)}
\begin{itemize}
    \item Total: 577 samples
    \begin{itemize}
        \item Against: 198 (34.3\%)
        \item Neutral: 208 (36.0\%)
        \item Pro: 171 (29.6\%)
    \end{itemize}
\end{itemize}

\textbf{Topic 2:~\textarabic{ وجود اللاجئين والمهاجرين في الأردن من فلسطين، سوريا، العراق} (Refugee/Immigrant Presence in Jordan)}

\begin{itemize}
    \item Total: 628 samples
    \begin{itemize}
        \item Against: 228 (36.3\%)
        \item Neutral: 163 (26.0\%)
        \item Pro: 237 (37.7\%)
    \end{itemize}
\end{itemize}

\textbf{Data Split}
\begin{itemize}
    \item Training: 843 samples (70\%)
    \item Development: 181 samples (15\%)
    \item Test: 181 samples (15\%)
\end{itemize}




\textbf{Dataset Examples}

\textbf{Topic: Normalization with Israel}

\begin{itemize}
    \item \textbf{Pro:} ``\textarabic{الجامعة العربية قالت إنها لا ترى أن \#التطبيع مع \#إسرائيل خطوة ضد القضية الفلسطينية}''
    
    \textit{English:} The Arab League said it does not see normalization with Israel as a step against the Palestinian cause.
    
    \item \textbf{Against:} ``\textarabic{إن قدرة الدولة على التطبيع جهارًا نهارًا مع النظام الإسرائيلي تتماشى مع قوة واستقرار نظامها المستبد}''
    
    \textit{English:} A state's ability to normalize openly with the Israeli regime aligns with the strength and stability of its authoritarian system.
    
    \item \textbf{Neutral:} ``\textarabic{كشف تقرير نشرته البوابة الإسرائيلية لشؤون الزراعة أنّه منذ سنة ونصف تجري في أوروبا اجتماعات بين ممثلين إسرائيليين واماراتيين}''
    
    \textit{English:} A report published by the Israeli agricultural portal revealed that meetings between Israeli and Emirati representatives have been taking place in Europe for a year and a half.
\end{itemize}

\textbf{Topic: Refugee/Immigrant Presence in Jordan}

\begin{itemize}
    \item \textbf{Pro:} ``~\textarabic{لا مكان للعنصرية بالأردن اي شخص داخل الأردن يعامل معاملة ابن البلد وهاذا الشخص يمثل كل أردني شريف}''
    
    \textit{English:} There is no place for racism in Jordan. Any person inside Jordan is treated like a son of the country, and this person represents every honorable Jordanian.
\end{itemize}

\subsection{Setup and Evaluation}
Subtask~B follows the same two-phase setup (development and test) and evaluation protocol as Subtask~A (see Section~\ref{sec:taskSetup}), with Macro F1 as the primary metric computed over the three labels: \textsc{Favor}, \textsc{Against}, and \textsc{Neither}.

\subsection{Results and Overview of the Systems}
A total of 8 teams submitted runs during the evaluation phase of Subtask~B (Cross-Topic Stance Detection), of which 6 teams submitted system descriptions and system papers. In Table~\ref{tab:subtaskB-systems}, we provide an overview of the participating systems for which a description was submitted. In Table~\ref{tab:subtaskB-results}, we report the official results for those teams, ranked by Macro F1.


As shown in Table~\ref{tab:subtaskB-systems}, fine-tuning Arabic pre-trained transformer models is the dominant strategy, with several teams incorporating cross-validation, ensemble methods, and topic-conditioned input formulations to improve cross-topic generalization.

\begin{table*}[t]
\centering
\small
\caption{Subtask B: Overview of the participating systems. DL: Deep Learning, ML: Classic Machine Learning.}
\label{tab:subtaskB-systems}
\resizebox{\textwidth}{!}{%
\begin{tabular}{lccccccccccc}
\toprule
\multirow{2}{*}{\textbf{Team}} &
\multicolumn{5}{c}{\textbf{Transformer}} &
\multicolumn{6}{c}{\textbf{Misc.}} \\
\cmidrule(lr){2-6}\cmidrule(lr){7-12}
 & \rotatebox{90}{AraBERT (Twitter)} & \rotatebox{90}{MARBERT} & \rotatebox{90}{XLM-RoBERTa} & \rotatebox{90}{mDeBERTa-v3} & \rotatebox{90}{CAMeL-BERT} & \rotatebox{90}{Data Augmentation} & \rotatebox{90}{Preprocessing} & \rotatebox{90}{Hyperparameter Tuning} & \rotatebox{90}{Cross-validation} & \rotatebox{90}{Ensemble Methods} & \rotatebox{90}{Topic Conditioning} \\
\midrule
Viva\_Palestine~\cite{vivapalestine_2026}          & \checkmark & \checkmark &           &           &           &           & \checkmark & \checkmark &           &           & \checkmark \\
EGCSS~\cite{egcss_2026}                    & \checkmark &           &           &           &           & \checkmark & \checkmark & \checkmark & \checkmark &           & \checkmark \\
U4RASD~\cite{u4rasd_2026}                   &           & \checkmark &           &           &           & \checkmark &           &           & \checkmark &           &           \\
The Resistant Word~\cite{resistantword_2026}       &           & \checkmark & \checkmark &           & \checkmark &           &           & \checkmark & \checkmark & \checkmark &           \\
A2NLP~\cite{a2nlp_2026}                    & \checkmark &           &           &           &           & \checkmark & \checkmark & \checkmark & \checkmark &           & \checkmark \\
The Blackwell Collective~\cite{blackwell_2026} &           &           &           & \checkmark &           &           & \checkmark & \checkmark &           &           & \checkmark \\
\bottomrule
\end{tabular}%
}
\end{table*}

\begin{table}[t]
\centering
\small
\caption{Official results for Subtask B, ranked by Macro F1 score. Leaderboard ranks are shown alongside overall submission ranks (in parentheses). Of the 8 total test submissions, only 6 teams provided a model description and are included here.}
\label{tab:subtaskB-results}
\begin{tabular}{lcc}
\toprule
\textbf{Team} & \textbf{Rank} & \textbf{Macro F1} \\
\midrule
Viva\_Palestine          & 1 (\#1) & 0.8724 \\
EGCSS                    & 2 (\#2) & 0.8607 \\
U4RASD                   & 3 (\#3)& 0.8601 \\
The Resistant Word       & 4 (\#4)& 0.8562 \\
A2NLP                    & 5 (\#5)& 0.8483 \\
The Blackwell Collective & 6 (\#7) & 0.7407 \\
\bottomrule
\end{tabular}
\end{table}

Team \textbf{Viva\_Palestine}~\cite{vivapalestine_2026} achieved the best performance in Subtask~B with a Macro F1 of 0.8724, ranking first on the leaderboard. They fine-tuned \texttt{UBC-NLP/MARBERT} by merging the training and validation sets to form an expanded training set of 1,024 records. The input was formatted as a topic--sentence pair truncated to 128 tokens. The best hyperparameters were: learning rate of $2\times10^{-5}$, batch size of 8, 4 training epochs, warmup ratio of 0.1, weight decay of 0.01, and the AdamW optimizer.

Team \textbf{EGCSS}~\cite{egcss_2026} ranked second with a Macro F1 of 0.8607. They fine-tuned \texttt{bert-base-arabertv02-twitter} with a classification head, incorporating topic descriptions concatenated to the input to provide cross-topic signal. 
They experimented with data augmentation via back-translation, Claude-generated tweets, and the ArabicStanceX dataset, but none of these yielded improvements over the base configuration and were excluded from the final submission. The best hyperparameters were: learning rate of $2\times10^{-5}$, maximum length of 64, 10 epochs with a linear learning rate schedule, and early stopping based on the labeled validation set.

Team \textbf{U4RASD}~\cite{u4rasd_2026} ranked third with a Macro F1 of 0.8601. 
Their final system fine-tuned MARBERTv2 with dialect-aware LLM-based data 
augmentation, using Gemini 3 Flash Preview to generate three paraphrased variants 
of each training sample while preserving both stance label and dialectal register. 
They also investigated contrastive learning, multi-task learning with LLM-generated 
auxiliary labels, counterfactual augmentation, and zero-shot prompting, none of 
which improved over the dialect-aware augmentation baseline. Final model selection 
was based on the best Macro F1 on the development set using stratified 5-fold 
cross-validation.

Team \textbf{The Resistant Word}~\cite{resistantword_2026} ranked fourth with a Macro F1 of 0.8562. They employed a multi-model ensemble with 5-fold stratified cross-validation, fine-tuning four pre-trained Arabic transformer models: MARBERT, AraBERT Large, XLM-RoBERTa Base, and CAMeLBERT-Mix. Each model received a topic-sentence pair truncated to 128 tokens. Prior to training, random oversampling was applied to balance the training set. All models were trained for 5 epochs with a learning rate of $2\times10^{-5}$, batch size of 16, and BF16 mixed precision. A custom weighted cross-entropy loss was applied, with class weights computed inversely proportional to class frequency. Final predictions were produced by averaging softmax probabilities across all 20 runs (4 models $\times$ 5 folds).

Team \textbf{A2NLP}~\cite{a2nlp_2026} ranked fifth with a Macro F1 of 0.8483. They fine-tuned \texttt{aubmindlab/bert-base-arabertv02-twitter} with a prompt-based input formulation that explicitly conditions stance prediction on the target topic.
Preprocessing was tailored to Arabic social media and included emoji normalization, URL and mention removal, diacritic removal, Alef variant unification, and whitespace normalization. Stratified 5-fold cross-validation was employed with model selection based on validation Macro F1. A weighted cross-entropy loss addressed class imbalance. The best hyperparameters were: learning rate of $2\times10^{-5}$, batch size of 16, up to 10 epochs with early stopping (patience = 2).

Team \textbf{The Blackwell Collective}~\cite{blackwell_2026} ranked sixth with a Macro F1 of 0.7407. They applied the same Statement Tuning architecture as in Subtask~A, fine-tuning \texttt{microsoft/mdeberta-v3-base} with cloze-style input templates and a \texttt{[MASK]}-based prototype classifier. Topic-Conditioned Layer Normalization (T-CLN) served as the primary mechanism for cross-topic generalization, dynamically conditioning feature distributions on a learnable 64-dimensional topic embedding. Prototype Contrastive Learning (PCL) and R-Drop regularization were also applied. Training followed the same two-stage schedule as in Subtask~A: backbone frozen for 2 epochs, then full fine-tuning for 8 additional epochs using AdamW.

\section{Conclusion and Future Work}
\label{sec:conclusion}
We presented an overview of the StanceNakba 2026 Shared Task, which addresses stance detection in polarized social media discourse on the Palestinian-Israeli conflict. The task introduced a dual-framework approach consisting of two subtasks: \textbf{Subtask~A} (Actor-Level Stance Detection), identifying whether authors express Pro-Palestine, Pro-Israel, or Neutral orientations; and \textbf{Subtask~B} (Cross-Topic Stance Detection), detecting Favor, Against, or Neither stances toward specific conflict-related topics, namely normalization with Israel and refugee presence in Jordan.

The task attracted considerable interest from the research community. A total of 7 teams made official submissions for Subtask~A, and 6 teams for Subtask~B, with two teams participating in both subtasks. For both subtasks, the majority of systems fine-tuned pre-trained Arabic transformer models, with BERT-based architectures — particularly MARBERT, AraBERT, and DeBERTa-v3 variants — being the most common backbone. Top-performing systems on Subtask~A achieved Macro F1 scores as high as 0.9620, while Subtask~B results were competitive with the best system reaching a Macro F1 of 0.8724, reflecting the added difficulty of cross-topic generalization. Several teams incorporated other techniques such as cross-validation, ensemble methods, topic-conditioned input formulations, and consistency regularization to improve performance.

Future editions of the StanceNakba shared task could explore more fine-grained stance categories and extend coverage to additional conflict-related topics and Arabic dialects. Increasing dataset size, particularly for underrepresented stance classes, and developing evaluation protocols that better capture cross-topic generalization will be important directions. 
Future work could also investigate the role of large language models (LLMs) under zero-shot and few-shot settings.

\section{Limitations}
\label{sec:limitations}
Several participating teams noted difficulty predicting the neutral or ``neither'' class, likely due to its inherent ambiguity. Additionally, the cross-topic nature of Subtask~B introduces domain shift between training and test topics, which current transformer-based models do not fully address. Expanding the dataset with more balanced label coverage and including a broader range of topics and dialectal varieties would help mitigate these limitations in future iterations of the task.

\section{Ethical Considerations \& Dataset Availability}
\label{sec:ethical}
\subsection{Data Source and Consent}

The datasets used in the StanceNakba Shared Task consist of publicly available posts collected from Twitter X in accordance with the platform's terms of service. Only publicly accessible content was included, and no attempt was made to access private accounts, restricted material, or deleted posts.

Usernames, profile information, and other direct identifiers were removed. The dataset does not include user metadata beyond the textual content necessary for stance classification. In compliance with platform policy, only post IDs and annotation labels will be released where required.

\subsection{Sensitive and Political Content}

This shared task focuses on discourse related to the Palestinian Israeli conflict and related regional issues. This topic is politically sensitive and may involve references to violence, displacement, discrimination, or traumatic events.

The dataset may contain polarized language and emotionally charged expressions. The purpose of the shared task is scientific analysis of stance detection and cross topic generalization. It is not intended to endorse, legitimize, or amplify any political position. Results should not be interpreted as normative judgments.

\subsection{Risk of Harm and Misuse}

Automated stance detection systems applied to political discourse may be misused for surveillance, political profiling, targeted persuasion, or repression. We explicitly discourage the use of models trained on this dataset for surveillance or discriminatory purposes.

Because Subtask A involves inference of general political alignment, predictions should not be treated as reliable indicators of an individual's beliefs or identity. Model outputs are probabilistic classifications based on limited textual evidence and may be incorrect.

\subsection{Annotation Bias and Subjectivity}

Stance annotation is inherently interpretive. Although detailed guidelines were developed for the Pro Palestine, Pro Israel, Neutral, Favor, Against, and Neither labels, annotators may bring implicit cultural, political, or linguistic biases.

To mitigate this risk, annotators were trained using standardized definitions, disagreements were adjudicated, and balanced label distributions were maintained where feasible. Nevertheless, residual bias may remain in both the English and Arabic datasets. Models trained on this data may learn annotation artifacts rather than purely linguistic signals of stance.

\subsection{Representation and Sampling Limitations}

The datasets consist of a limited number of annotated English and Arabic social media posts. They are not representative of all political perspectives, demographic groups, or geographic populations. Social media users are themselves not representative of broader societies.

Accordingly, findings should not be generalized to entire populations or interpreted as reflecting majority opinion. Claims about cross topic generalization should be framed cautiously.

\subsection{Cross Lingual and Cross Cultural Considerations}

Subtask A and Subtask B involve different languages and sociopolitical contexts. Linguistic markers of stance may vary across dialects, regions, and political cultures. Models may encode unintended cultural assumptions.

Participants are encouraged to report per class performance, conduct qualitative error analysis, and critically examine model behavior across topics and languages.

\subsection{Annotator Well Being}

Exposure to polarized or conflict related discourse may cause emotional fatigue. Annotation processes should include reasonable workload distribution and allow annotators to opt out of reviewing content they find distressing.

\subsection{Data Availability}

The dataset is released for non commercial research purposes. Users must comply with platform terms of service, avoid attempts at deanonymization, and refrain from surveillance or discriminatory applications. The dataset can be accessed upon request through the following form:
\url{https://forms.gle/YUFdA16R6HkSZjp88}

\section*{Acknowledgment}
This shared task was made possible by the National Priorities Research Program grant NPRP14C-0916-210015 from the Qatar National Research Fund (QNRF), part of the Qatar Research, Development and Innovation Council (QRDI).

\bibliographystyle{lrec2026-natbib}
\bibliography{References_main}


\end{document}